\documentclass[conference]{IEEEtran}
\IEEEoverridecommandlockouts
\usepackage{url}
\usepackage{array,booktabs,ragged2e}
\usepackage{subcaption}
\usepackage{tabularx}
\usepackage{cite}
\usepackage{amsmath,amssymb,amsfonts}
\usepackage{algorithmic}
\usepackage{graphicx}
\usepackage{textcomp}
\usepackage{xcolor}
\usepackage{microtype}
\def\BibTeX{{\rm B\kern-.05em{\sc i\kern-.025em b}\kern-.08em
    T\kern-.1667em\lower.7ex\hbox{E}\kern-.125emX}}
\begin{document}

\title{Robust Summarization of Doctor-Patient Conversations: TalTech Systems for the Beyond Transcription Challenge}

\author{\IEEEauthorblockN{Aivo Olev}
\IEEEauthorblockA{\textit{Tallinn University of Technology}\\
Tallinn, Estonia \\
aivo.olev@taltech.ee}
\and
\IEEEauthorblockN{Tanel Alum\"ae}
\IEEEauthorblockA{\textit{Tallinn University of Technology}\\
Tallinn, Estonia \\
tanel.alumae@taltech.ee}
}

\maketitle

\begin{abstract}

This paper describes TalTech's submissions to the Beyond Transcription Challenge (BeTraC), which requires generating SOAP notes directly from long doctor-patient conversation recordings, without intermediate transcription.
After screening open-weight speech LLMs for long-audio robustness, we adapted Voxtral Mini (lightweight track) and Voxtral Small (heavyweight track) with LoRA supervised fine-tuning followed by DAPO reinforcement learning that uses the challenge metric, Open Medical Concept F1, as its reward.
Our systems ranked first in both tracks, and an independent LLM-as-a-judge evaluation showed the lowest hallucination rate among all submissions, indicating that reinforcement learning against a concept-matching metric need not compromise factual reliability.
We also find that fine-tuning on text transcripts transfers well to speech input and appears to improve robustness on out-of-domain real recordings.

\end{abstract}


\section{Introduction}

Clinical note generation from doctor-patient dialogue recordings is attractive because it can reduce documentation burden without requiring a separate transcription step.
However, such systems must accurately preserve clinical facts, because hallucinated symptoms, diagnoses, or treatment plans can render an otherwise well-written note unsafe.
The BeTraC challenge targets this setting by requiring systems to generate plain-text medical SOAP (subjective, objective, assessment, and plan) notes from audio while disallowing intermediate transcription as an explicit system component \cite{betrac}.

Team TalTech submitted direct audio-to-SOAP systems to both BeTraC tracks.
We built around Voxtral, a family of open-weight speech understanding models with strong audio and text capabilities~\cite{liu2025voxtral}, selected after a zero-shot study on the validation split in which Voxtral showed the best speech-to-summary and speech recognition behavior of the speech LLMs tested. We then adapted Voxtral Mini for the lightweight track and Voxtral Small for the heavyweight track using supervised fine-tuning (SFT) followed by DAPO reinforcement learning.

\section{Challenge, Data, and Metrics}

BeTraC uses the SynthDoPaCo corpus, a fully synthetic dataset of two-speaker doctor-patient conversations with ambient noise, room reverberation, and compression artifacts~\cite{labrak2026generating}. It contains 8,800 conversations, approximately 1,329 hours of audio, about nine minutes per conversation, split into 7,200 training and 400 validation dialogues.
The final leaderboard is based on a blind testset of 875 audio examples.

Both tracks require open-weight models and prohibit intermediate transcripts; the lightweight track limits total model size to 6B parameters and disallows tool use or agentic pipelines, while the heavyweight track allows up to 36B parameters.
Submissions are ranked by Open Medical Concept F1, computed with MeSH keyword matching and scispaCy-based named entity recognition; ROUGE-2 and ROUGE-3 F1 are secondary metrics \cite{betrac-eval}.

\section{Selection of Base Models}

Before fine-tuning, we compared several open-weight speech-capable LLMs on zero-shot speech-to-SOAP generation and speech recognition, asking whether a model can generate the required SOAP-note format from audio alone and process long audio robustly enough to support task-specific fine-tuning.
Table~\ref{tab:zeroshot} contains the results. We also considered Phi-4-multimodal-instruct (rejected for high GPU VRAM requirements on long audio) and Gemma-4-E2B/E4B (limited to 30-second audio inputs).

\newcolumntype{R}[1]{>{\RaggedLeft\arraybackslash}p{#1}}

\begin{table}[t]
\centering
\footnotesize
\caption{Zero-shot summarization and ASR results of audio LLMs.}
\label{tab:zeroshot}
\begin{tabular}{lrrR{0.9cm}R{0.9cm}}
\toprule
Model & C-F1$\uparrow$ & R-2$\uparrow$ & Mean WER$\downarrow$ & Median WER$\downarrow$ \\
\midrule
\multicolumn{2}{l}{\textit{\textbf{Lightweight models}}} \\
\midrule
Qwen2.5-Omni-3B \cite{Qwen2.5-Omni} & 0.260 & 0.092 & 39.1 & 43.3 \\
MOSS-Audio-4B-Instruct \cite{yang2026mossaudiotechnicalreport} & 0.314 & \textbf{0.140} & 10.0 & 2.3 \\
Voxtral-Mini-3B-2507 \cite{liu2025voxtral} & \textbf{0.354} & 0.121 & \textbf{2.5} & \textbf{1.3} \\
\midrule
\multicolumn{2}{l}{\textit{\textbf{Heavyweight models}}} \\
\midrule
Qwen2.5-Omni-7B \cite{Qwen2.5-Omni}  & 0.264 & 0.107 & 41.3 & 43.6 \\
Voxtral-Small-24B-2507 \cite{liu2025voxtral} & \textbf{0.384} & \textbf{0.196} & \textbf{2.8} & \textbf{0.9} \\
\bottomrule
\end{tabular}
\end{table}

Although BeTraC is not an ASR challenge, transcription quality was a useful diagnostic: a model that cannot reliably follow the long clinical conversation is unlikely to produce faithful SOAP notes.
Voxtral's low validation WER and stronger zero-shot SOAP outputs made it a clear choice for both tracks.
WER analysis also explains the other models' weak SOAP generation: Qwen2.5-Omni transcribes only roughly the first five minutes of audio, while MOSS-Audio-4B-Instruct's low median but high mean WER reflects hallucination loops on some inputs. Voxtral's long-audio robustness likely stems from its internal chunking of input into 30-second segments, which bounds GPU VRAM use and allows reusing Whisper's encoder without context-lengthening training.

\section{System Description}

\subsection{Supervised Fine-Tuning}

In the SFT phase, the base model is finetuned on the provided training set with cross-entropy loss. Voxtral-Mini was finetuned on speech inputs and Voxtral-Small on normalized dialogue transcripts, with speaker turn prefixes and non-speech event markers such as ``[sigh]'' removed.
Finetuning used LoRA with the audio encoder and audio projector of the base Voxtral model frozen, allowing both models to be trained on a single Nvidia H200 GPU (141 GB VRAM). Training used the ms-swift toolkit \cite{zhao2024swiftascalablelightweightinfrastructure} via a custom Voxtral plugin; hyperparameters are listed in Table \ref{tab:sft-hparams}. After training, the LoRA adaptors were merged into the model.

\begin{table}[t]
\centering
\scriptsize
\caption{Training hyperparameters.}
\label{tab:training-hparams}

\begin{subtable}[t]{0.49\linewidth}
\centering
\caption{Supervised finetuning.}
\label{tab:sft-hparams}
\begin{tabularx}{\linewidth}{@{}Xr@{}}
\toprule
Hyperparameter & Value \\
\midrule
LoRA target modules & all linear \\
LoRA rank / $\alpha$ / dropout & 16 / 16 / 0.05 \\
Effective batch size & 16 \\
Training epochs & 1.0 \\
Optimizer & fused AdamW \\
Learning rate & $1\times10^{-5}$ \\
LR scheduler / warmup  & cosine / 0.03 \\
Weight decay & 0.1 \\
Max gradient norm & 1.0 \\
\bottomrule
\end{tabularx}
\end{subtable}%
\hfill
\begin{subtable}[t]{0.42\linewidth}
\centering
\caption{Reinforcement learning.}
\label{tab:dapo-hparams}
\begin{tabularx}{\linewidth}{@{}Xr@{}}
\toprule
Hyperparameter & Value \\
\midrule
Loss type & DAPO \\
Reward & Concept F1 \\
Rollouts per sample & 8 \\
Steps per generation & 4 \\
KL $\beta$ & 0 \\
\bottomrule
\end{tabularx}
\end{subtable}

\end{table}

\subsection{Reinforcement Learning}

After SFT, we applied reinforcement learning with Open Medical Concept F1 as the reward and DAPO as the loss type. DAPO is a GRPO-family \cite{shao2024deepseekmath} refinement that improves learning efficiency and stability for long generated responses through token-level loss aggregation and modified clipping~\cite{yu2026dapo}. This suits SOAP notes, which are long enough that response-level rewards can otherwise be diluted across many tokens.
Table~\ref{tab:dapo-hparams} lists the RL-specific hyperparameters; the others match SFT.

\subsection{Contrastive Facts-and-SOAP model}
\label{ssec:facts-soap}

We also explored a fact-table-augmented variant of the lightweight model. For each training dialogue, DeepSeek-V4-Flash generated a structured table of clinically relevant facts supported by short transcript excerpts, each assigned to a SOAP-related category.
SFT targets were then formed by concatenating the fact table with the reference SOAP note, so the fact table acts as an intermediate reasoning trace emitted by the model itself rather than a pipeline component.
In the subsequent RL stage, the reward was computed only on the SOAP-summary portion, so the fact table shapes the model's internal organization during supervised learning while the RL objective stays aligned with the official scoring target.

\subsection{Results on validation data}

Table~\ref{tab:dev-results} lists model performance after the main fine-tuning stages, comparing speech inputs to the corresponding textual transcripts.
We draw three conclusions.
First, RL gives consistent improvements over SFT, even though both stages use the same training examples.
Second, summarization capability learned from text inputs transfers to speech input with only a small decrease in performance, consistent with earlier findings on cross-modal transfer in speech LLMs~\cite{Rouditchenko2025, choi2026exploring}.
Third, speech-input fine-tuning adds only a small further gain over text-input fine-tuning, suggesting it can be omitted for larger models when training cost is the main constraint.

\begin{table}[tb]
\centering
\footnotesize
\caption{Results on validation data after finetuning. Primary submissions in \textbf{bold}; contrastive lightweight model in \textit{\textbf{italic bold}}.}
\label{tab:dev-results}
\addtolength{\tabcolsep}{-0.2em}
\begin{tabular}{lcrrr}
\toprule
Model                                              & Input type & C-F1 & R-2 & R-3 \\ 
\midrule
\textit{Voxtral Mini} \\
\midrule
SFT transcript$\rightarrow$SOAP    & Transcript & 0.507 &	0.342 &	0.229     \\
SFT+RL transcript$\rightarrow$SOAP & Transcript & 0.555	& 0.371 &	0.253     \\
SFT transcript$\rightarrow$SOAP    & Audio      & 0.486 &	0.320 &	0.210     \\
SFT+RL transcript$\rightarrow$SOAP & Audio      & 0.544 &	0.359 &	0.242     \\ 
\textbf{SFT audio$\rightarrow$SOAP}    & Audio      &  0.496 &	0.340 & 	0.229     \\
SFT+RL audio$\rightarrow$SOAP & Audio      &  0.547 &	0.364 &	0.247     \\ 
\textbf{\textit{SFT+RL transcript$\rightarrow$Facts+SOAP}} & Audio      & 0.540 & 0.362 &	0.247    \\
\midrule
\textit{Voxtral Small} \\
\midrule
SFT transcript$\rightarrow$SOAP    & Transcript &  0.537 &	0.382 &	0.268     \\
SFT+RL transcript$\rightarrow$SOAP & Transcript &  0.576 &	0.409 &	0.289 \\
\textbf{SFT+RL transcript$\rightarrow$SOAP} & Audio      & 0.571 &	0.401 &	0.282    \\ 
\bottomrule
\end{tabular}
\end{table}

Based on these results, we selected the SFT+RL audio$\rightarrow$SOAP model as the lightweight primary submission and the SFT+RL transcript$\rightarrow$SOAP Voxtral Small model, applied to audio, as the heavyweight primary; the lightweight contrastive submission uses the facts-and-SOAP model.

\section{Results and Discussion}

\subsection{Official test-set results}

Table~\ref{tab:test-results} summarizes the official automated evaluation, computed by the organizers against reference notes generated by a Kimi-K2-based system from the gold dialogue transcripts~\cite{labrak2026generating}.
The blind test set combines the SynthDoPaCo test split (EE, 600 synthetic dialogues), acted patient-physician interviews from \cite{Fareez2022} (mock, 272 scored dialogues), and three realistic dialogues recorded for the challenge; submissions were ranked by Concept F1 on EE.

\begin{table}[t]
\centering
\scriptsize
\caption{Official test-set results (top-3 teams per track by the ranking metric, C-F1 on EE with 600 dialogues, plus our contrastive submission and the organizers' baseline). Mock has 272 scored dialogues, Real.\ only 3; R-2 and mean note length cover all 875 dialogues.}
\label{tab:test-results}
\setlength{\tabcolsep}{4pt}
\begin{tabular}{lrrrrrr}
\toprule
 & \multicolumn{4}{c}{C-F1$\uparrow$} & & \\
\cmidrule(lr){2-5}
Team & EE & Mock & Real. & All & R-2$\uparrow$ & Words \\
\midrule
\multicolumn{7}{l}{\textit{\textbf{Lightweight track}}} \\
\midrule
\textbf{TalTech (ours)} & \textbf{0.543} & 0.505 & 0.534 & \textbf{0.531} & 0.351 & 297 \\
NTT-HI-CS & 0.540 & \textbf{0.511} & 0.487 & \textbf{0.531} & \textbf{0.370} & 290 \\
\textit{TalTech (contrast)} & \textit{0.526} & \textit{0.510} & \textit{0.549} & \textit{0.521} & \textit{0.339} & \textit{306} \\
KUSLP & 0.515 & 0.465 & 0.431 & 0.499 & 0.355 & 297 \\
\midrule
\multicolumn{7}{l}{\textit{\textbf{Heavyweight track}}} \\
\midrule
\textbf{TalTech (ours)} & \textbf{0.563} & \textbf{0.555} & \textbf{0.607} & \textbf{0.560} & \textbf{0.395} & 295 \\
KUSLP & 0.544 & 0.530 & 0.516 & 0.540 & 0.387 & 291 \\
NTT-HI-CS & 0.539 & 0.517 & 0.510 & 0.532 & 0.368 & 294 \\
\midrule
Baseline (Qwen2.5-Omni-3B) & 0.245 & 0.254 & 0.244 & 0.248 & 0.080 & 409 \\
\bottomrule
\end{tabular}
\end{table}

Our primary submissions obtained the highest Concept F1 in both tracks (0.563 heavyweight, 0.543 lightweight).
The lightweight margin over NTT-HI-CS is narrow (0.543 vs.\ 0.540), and NTT-HI-CS achieves a higher ROUGE-2, an expected side effect of our RL stage, which optimizes Concept F1 rather than $n$-gram overlap.
Both systems more than double the baseline's Concept F1, consistent with our zero-shot screening in Table~\ref{tab:zeroshot}.

Two observations stand out.
First, test-set results transfer almost perfectly from validation. The heavyweight system scores 0.571 C-F1 / 0.401 R-2 on validation audio and 0.563 / 0.401 on EE, the lightweight system 0.547 / 0.364 versus 0.543 / 0.366; there is essentially no generalization gap, despite using the validation split for model, checkpoint, and reward selection.
Second, model scale was not decisive, as our Voxtral Mini-based lightweight system (0.543) effectively matches the second-best heavyweight submission (0.544), and the gain from Voxtral Mini to Voxtral Small within our recipe (+0.02 C-F1) is smaller than the gain from RL itself (+0.04--0.05 on validation).

\subsection{Robustness across test domains}

The three test subsets probe different distances from the synthetic training distribution: EE matches it, mock contains real acted recordings, and the realistic subset genuine spontaneous ones.
Our heavyweight system degrades least among the top systems when moving from synthetic to acted audio, losing 0.008 C-F1 (0.563$\rightarrow$0.555) versus 0.038 for our own lightweight primary.
A plausible contributing factor is that this model was fine-tuned on normalized \emph{transcripts}, with audio entering only at inference through the frozen Whisper-initialized encoder, so fine-tuning could not overfit to the synthetic TTS training audio; the audio-fine-tuned lightweight primary lacks this protection (though model capacity is a confound).
The facts-and-SOAP contrastive submission (Section~\ref{ssec:facts-soap}) also holds up better under domain shift: it trails the lightweight primary in-domain (0.526 vs.\ 0.543 on EE) yet matches it on mock (0.510 vs.\ 0.505) and scores higher on the realistic subset (0.549); its intermediate fact-extraction step appears likewise less tied to the synthetic domain.
On the realistic subset the heavyweight system reaches 0.607, although with only three dialogues this is anecdotal.

\subsection{LLM-as-judge evaluation}

\begin{table}[t]
\centering
\scriptsize
\caption{LLM-as-judge scores on EE (scale 1--5, higher better; hallucination/contradiction rates are fractions of atomic claims, lower better) for top-3 systems per track plus our contrastive submission.}
\label{tab:judge}
\begin{tabular}{lrrrrr}
\toprule
Team & Faith. & Cov. & Conc. & Halluc. & Contr. \\
\midrule
\multicolumn{6}{l}{\textit{\textbf{Lightweight track}}} \\
\midrule
\textbf{TalTech (ours)} & \textbf{4.71} & \textbf{4.49} & 4.16 & 0.25\% & \textbf{0.27\%} \\
NTT-HI-CS & 4.58 & 4.37 & 4.37 & \textbf{0.24\%} & 0.55\% \\
\textit{TalTech (contrast)} & \textit{4.55} & \textit{4.42} & \textit{4.22} & \textit{0.39\%} & \textit{0.40\%} \\
KUSLP & 4.59 & 4.46 & 4.11 & 0.38\% & 0.37\% \\
\midrule
\multicolumn{6}{l}{\textit{\textbf{Heavyweight track}}} \\
\midrule
\textbf{TalTech (ours)} & 4.80 & \textbf{4.50} & \textbf{4.41} & \textbf{0.08\%} & 0.16\% \\
KUSLP & \textbf{4.81} & 4.32 & 4.34 & 0.17\% & \textbf{0.10\%} \\
NTT-HI-CS & 4.47 & 4.38 & 4.30 & 0.41\% & 0.30\% \\
\bottomrule
\end{tabular}
\end{table}

The organizers additionally scored all submissions with an LLM-as-judge pipeline \cite{labrak2026generating}: each note is decomposed into atomic claims that a Gemma~4 31B judge verifies against the gold transcript rather than the Kimi-K2 reference notes, making it partially independent of the automated metrics; Table~\ref{tab:judge} shows results on the ranking subset.

Our heavyweight system obtains the lowest hallucination rate of any scored submission (0.08\% on EE), the best coverage and conciseness in its track, and faithfulness indistinguishable from the best (4.80 vs.\ 4.81); our lightweight primary achieves the highest faithfulness (4.71) and coverage (4.49) in its track.

This addresses the main risk of our recipe. Optimizing Concept F1 directly with RL invites a degenerate strategy of enumerating plausible medical concepts to inflate recall, yielding long, noisy notes.
The judge scores show this did not happen. Our notes remain average-length (295--297 words), concept precision stays \emph{above} recall (0.59 vs.\ 0.54, heavyweight), and hallucination rates stay low; the F1 reward's precision term suffices to prevent concept stuffing.

Finally, since the ranking metric matches keywords and entities against LLM-generated references, one may ask whether it measures note quality or mere surface overlap.
Across the 21 judged submissions, system-level Concept F1 on EE correlates strongly with judge faithfulness (Spearman $\rho=0.91$) and inversely with hallucination rate ($\rho=-0.93$), supporting the metric choice, though both signals ultimately derive from LLMs and per-note agreement may be weaker.

\section{Conclusion}

We presented TalTech's winning systems for both BeTraC tracks: Voxtral models adapted to direct audio-to-SOAP generation with LoRA SFT and DAPO RL against the challenge metric.
Zero-shot screening for long-audio robustness proved an effective model-selection criterion; RL added 0.04--0.05 Concept F1 over SFT without inflating note length or hallucinations; transcript-based fine-tuning transferred to speech input and appears to help under domain shift; model scale was secondary.
Future work includes validation on real clinical recordings, mixed transcript-audio supervision, and faithfulness-aware rewards.

\section*{Acknowledgment}

Generative AI tools (Anthropic Claude Fable and OpenAI GPT-5.5) were used for grammar and style correction of the manuscript text. AI was not used to write any significant part of the paper.

\bibliographystyle{IEEEbib}
\bibliography{refs}

\end{document}